\documentclass[letterpaper, 10 pt, conference]{ieeeconf}  %

\IEEEoverridecommandlockouts                              %

\usepackage{graphicx} %
\usepackage{epsfig} %
\usepackage{epigraph}
\usepackage{url}
\usepackage{textgreek}
\usepackage{todonotes}
\usepackage{wrapfig}

\title{\LARGE \bf
 REPLAB: A Reproducible Low-Cost Arm Benchmark Platform for Robotic Learning
}

\author{Brian Yang$^{1}$, Jesse Zhang$^{2}$, Vitchyr Pong$^{3}$, Sergey Levine$^{4}$, and Dinesh Jayaraman$^{5}$%
\thanks{*This research was supported by Berkeley DeepDrive, NSF IIS-1651843, and the Office of Naval Research (ONR), Google, NVIDIA, and Amazon.}%
\thanks{$^{1}$ brianhyang@berkeley.edu, UC Berkeley.}%
\thanks{$^{2}$ jessezhang@berkeley.edu, UC Berkeley.}%
\thanks{$^{3}$ vitchyr@berkeley.edu, UC Berkeley.}%
\thanks{$^{4}$ svlevine@eecs.berkeley.edu, UC Berkeley.}%
\thanks{$^{5}$ dineshjayaraman@eecs.berkeley.edu, UC Berkeley.}%
}

\begin{document}

\maketitle
\thispagestyle{empty}
\pagestyle{empty}

\begin{abstract}
  Standardized evaluation measures have aided in the progress of machine learning approaches in disciplines such as computer vision and machine translation. In this paper, we make the case that robotic learning would also benefit from benchmarking, and present the ``REPLAB'' platform for benchmarking vision-based manipulation tasks. REPLAB is a reproducible and self-contained hardware stack (robot arm, camera, and workspace) that costs about 2000 USD, occupies a cuboid of size 70x40x60 cm, and permits full assembly within a few hours. Through this low-cost, compact design, REPLAB aims to drive wide participation by lowering the barrier to entry into robotics and to enable easy scaling to many robots. We envision REPLAB as a framework for reproducible research across manipulation tasks, and as a step in this direction, we define a template for a grasping benchmark consisting of a task definition, evaluation protocol, performance measures, and a dataset of 92k grasp attempts. We implement, evaluate, and analyze several previously proposed grasping approaches to establish baselines for this benchmark. Finally, we also implement and evaluate a deep reinforcement learning approach for 3D reaching tasks on our REPLAB platform. Project page with assembly instructions, code, and videos: \url{https://goo.gl/5F9dP4}.
\end{abstract}

\section{Introduction}

Since the 90's, the study of artificial intelligence has been transformed by data-driven machine learning approaches. This has been accompanied and enabled by increased emphasis on reproducible performance measures in fields like computer vision and natural language processing. While benchmark-driven research has its pitfalls~\cite{muller2018measuring,torralba2011unbiased}, well-designed benchmarks and datasets~\cite{deng2009imagenet,zue1990speech,lewis2004rcv1} drive increased research focus on important problems, provide a way to chart the progress of a research community, and help to quickly identify, disseminate, and improve upon ideas that work well.

In robotic manipulation, establishing effective benchmarks has proven exceedingly challenging, especially for robotic learning, where the principal concern is with the generalization of learned models to new objects and situations, rather than raw proficiency on a single narrow task. An important reason for this is that progress in robotics comes not only through improvements in control algorithms, but also through improvements in hardware (such as sensing and actuation). Traditional approaches to robotic control are closely intertwined with the specifics of the robotic hardware---for instance, grasping with a parallel-jawed gripper, a five-fingered hand, and a suction cup would all be treated as different tasks, each requiring their own different control algorithms. In this view, the large space of hardware choices and tasks makes it futile to attempt to meaningfully measure progress through a few focused benchmarks.

\begin{figure}[t]
  \centering
  \includegraphics[width=1\linewidth]{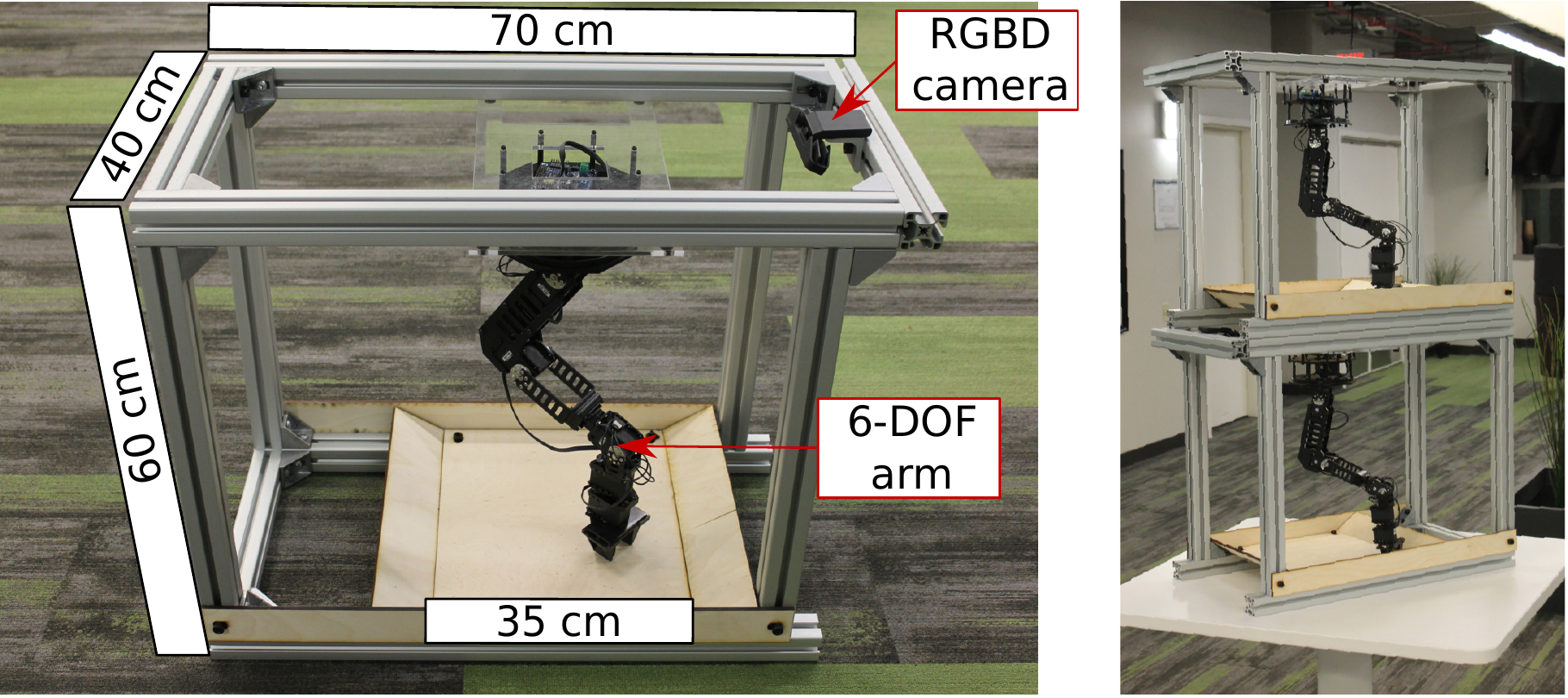}
  \caption{\textbf{(Left)} One REPLAB cell with annotated dimensions \textbf{(Right)} Two REPLAB cells stacked on top of each other on a desk.}
  \label{fig:replab}
\end{figure}

However, in the light of relatively recent changes in the research landscape, we contend that it may now be time to reconsider the idea of manipulation benchmarks. First, research in machine learning-based  manipulation~\cite{lenz2015deep,levine2018learning,pinto2016supersizing,gualtieri2016high,viereck2017learning,mahler2017dex,calandra2017feeling,zeng2017robotic,calandra2018more,deisenroth2011learning,gupta2018robot} aims to develop data-driven approaches that are, at least to some degree, agnostic to the particular choice of hardware---although models trained on one platform are unlikely to work on another, the same learning algorithms can in principle be deployed on new platforms with minimal additional engineering. The performance of such an approach on one hardware platform is generally expected to be predictive of its performance on other platforms too. Given this, we might hope that progress in learning-based control may be treated as orthogonal to hardware improvements.  Thus it may now be possible to meaningfully consolidate the space of task definitions and hardware configurations to a small representative set, which is a prerequisite for defining a benchmark.

Today's robotics hardware is already  mature enough to permit the human-teleoperated performance of tasks that are substantially harder than those that can be done with automated control methods~\cite{vertut2013teleoperation}. It is therefore reasonable to conclude that control, not hardware, is now the primary bottleneck for progress in robotics, and manipulation in particular. This means that a robotic learning benchmark is not merely \emph{possible} as discussed above, it could potentially serve a very important purpose to the research community.

What would a manipulation benchmark accomplish? Recent reinforcement learning (RL) benchmarks such as ALE~\cite{bellemare2013arcade} and Open AI Gym~\cite{gym} are useful reference points to answer this question. They serve three key functions for the RL community: enabling apples-to-apples comparison of RL algorithms by standardizing environments and tasks, enabling fast and easy replication and improvement of research ideas, and driving increased research by lowering the barrier to entry into RL.

In this paper, we remove one important hurdle for a manipulation benchmark by proposing a standardized and easily replicable hardware platform: we describe a reproducible ``REPLAB'' work cell design based on a low-cost commercially available robotic arm and an RGBD camera. In addition, we provide a template for a benchmark based on this platform, focusing on arguably the most widely studied robotic manipulation task: grasping. We present a dataset that can be used together with REPLAB to evaluate learning algorithms for robotic grasping. We implement and evaluate prior grasping approaches on this cell platform to serve as a template for a REPLAB benchmark.

REPLAB's design is motivated by the following goals, in order of priority: (i) to facilitate consistent and reproducible progress metrics for robotic learning,
(ii) to lower the barrier to entry into robotics for researchers in related disciplines like machine learning so that robotic learning research is not restricted only to a small number of well-established labs,
(iii) to encourage and enable plug-and-play reproducible software implementations of robotic learning algorithms, by promoting a standardized and exhaustively specified platform,
(iv) to allow  easily scaling up and parallelizing robotic learning algorithms across multiple robots, and promote testing for generalization to new robots,
(v) to facilitate crowdsourcing data collection efforts across REPLAB cells potentially distributed across multiple research labs, and
(vi) to be able to afford to evolve through iterative community-driven improvement of the REPLAB platform itself, a luxury that would not be available with a more expensive design.
\section{Related Work}\label{sec:related}

Robotic approaches today are largely tested in custom settings: environments, hardware, task definitions, and performance measures may all vary from paper to paper. %
While the problem this raises for measuring the effectiveness of different approaches is widely acknowledged in the robotics community~\cite{bonsignorio2015toward,del2006benchmarks,dillmann2004ka,muller2018measuring}, solutions have been elusive.

The majority of prior approaches to benchmarking in robotics have taken the form of a live competition between approaches, e.g., the DARPA Grand Challenge, Amazon Picking Challenge, and RoboCup. Each competing approach, consisting of specific hardware setups as well as control algorithms, is tested in the same physical location. This provides valuable performance measures of complete robotic systems, but it is logistically difficult to provide more than sparse point estimates of performance for each approach on a yearly basis. %

Beyond such live competitions, for grasping in particular, there have been other previous efforts to standardize various aspects of the task. The YCB dataset~\cite{calli2015benchmarking} is a standardized diverse object set for evaluating all grasping approaches. The ACRV benchmark~\cite{leitner2017acrv} goes one step further and proposes not only a standard object set, but also a standard test setting with a specified shelf design and specified object configurations within the shelf. The authors of DexNet~\cite{mahler2017dex} share a dataset of synthetic grasps to train grasp quality convolutional networks, and offer to perform on-robot evaluation of models with high accuracy on held-out grasps. OpenGRASP\cite{ulbrich2011opengrasp} proposes fully standardized task, hardware, and performance metrics are for grasping, but in a simulated environment. %
To our knowledge, ours is the first effort to propose a benchmark framework consisting of a standardized \emph{complete} stack for real-world grasping, including the full hardware configuration (such as robot, sensors, and work cell design), dataset of real-world grasps, software implementations of baselines, and performance measures.

REPLAB is also built with collective robotics in mind. Prior efforts in this direction include~\cite{levine2018learning,kalashnikov2018qt}, where data collection for grasping was parallelized across many physically collocated robots. Rather than a such a collocated group, the Million Object Challenge (MOC)~\cite{tellex} aims to crowdsource grasping data collection from 300 Baxter robots all over the world. REPLAB cells are designed to fit both these cases, since they are low-cost, low-volume, reproducible, and stackable: 20 REPLAB cells stacked to about 2m elevation occupy about the same floor space and cost less than two times as much as a single Baxter arm. The closest effort to this~\cite{gupta2018robot} trains grasping policies for low-cost mobile manipulators by collecting data from several such manipulators under varying lighting conditions.  %

Finally, previous efforts have also provided standardized and easily accessible full hardware stacks such as Duckietown for navigation~\cite{paull2017duckietown} and Robotarium for swarm robotics~\cite{pickem2017robotarium}. We share their motivation of democratizing robotics and driving increased participation, and our focus is on manipulation tasks.

\section{REPLAB Cell Design Overview}

We now describe various key aspects of the design of the REPLAB platform. Exhaustive documentation for constructing a complete REPLAB cell is hosted at: \url{https://goo.gl/5F9dP4}.

\subsection{Cell Design}

A REPLAB cell, shown in Fig~\ref{fig:replab}, is a portable, self-contained complete hardware stack (arm, sensor, workspace, and cage) for manipulation tasks. It occupies a cuboid of size 70x40x60 cm (length, width, height). The outer cage is constructed with easily composable lightweight 80-20 rods manufactured to our specifications. A low-cost  WidowX arm from Interbotix Labs is suspended upside down and its base is mounted to the ceiling of the cell to maximize its reachable effective workspace. The arm has six degrees of freedom: a 1-DOF rotating base, three 1-DOF joints, and a 1-DOF rotating wrist, and a 1-DOF parallel-jawed gripper with minimum width 1 cm and maximum width 3 cm.
A Creative Blasterx Senz3D SR 300 RGB-Depth camera is mounted to a specified standard position on the ceiling near the front of the cell so that the entire workspace is comfortably within its optimal field of view and operating distance. Mounts for the robot, the camera, and a 35x40 cm workspace are manufactured through laser cutting. A full list of parts, laser cutting templates, and all design parameters are exhaustively recorded shared on the project page for reproducibility. We verified that an undergraduate student with little prior robotics experience was able to build a REPLAB cell within three hours, given a pre-assembled arm, other required components, and our assembly instructions. %

The physical cell dimensions are designed to allow stacking of multiple cells on top of one another, as shown in Fig~\ref{fig:replab}. With our current design, we expect that it will be feasible for up to 20 arm cells, stacked to 2.2 metres in height (four cell heights), to occupy the same floor space as a typical setup for a single Baxter arm, for instance.

\begin{figure*}[t]
  \centering
  \includegraphics[width=\linewidth]{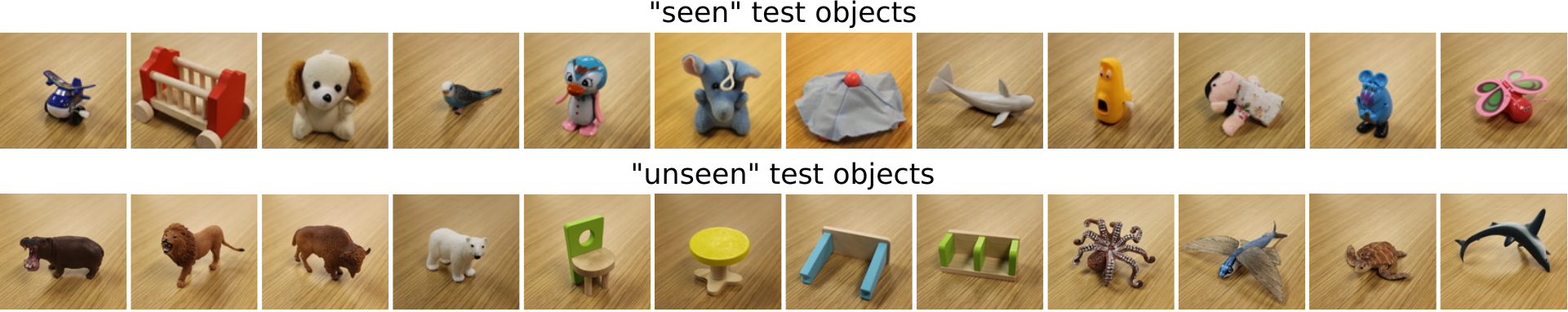}
  \caption{We train learning-based grasping approaches on over 50k grasp attempts with over 100 objects, and evaluate them on two sets of objects: 20 seen objects (sampled from the training objects), and 20 unseen objects from a different distribution with more complicated shapes. Here, we show a subset of seen (top) and unseen (bottom) test objects used in our benchmark evaluations.}
  \label{fig:objects}
\end{figure*}

A single REPLAB cell costs about 2000 USD, and can be assembled in a few hours. Together with one spare servo for each of six servos on the arm, the cost is under 3000 USD. This is comparable to the cost of a single workstation. During experiments for this paper, REPLAB cells proved to be quite robust. With software constraints in place to avoid arm collisions with the boundaries of the REPLAB cell, we encountered no major breakages over more than 100,000 grasp attempts. No servos needed to be replaced. Repair maintenance work was largely limited to occasional tightening of screws and replacing frayed cables.

\subsection{Camera-Arm Calibration}~\label{sec:calibration}
We perform camera-robot calibration on a single REPLAB cell by using a checkerboard and registering robot coordinate points to 3D image pixel coordinates (x, y, depth) from the camera. Since the cell design is exhaustively specified, our construction protocol ensures that the same calibration matrix may be reused for other cells.
\begin{figure}
  \begin{tabular}{@{}cc@{}}
       \small{uncalibrated} & \small{calibrated} \\
       \includegraphics[width=0.47\linewidth]{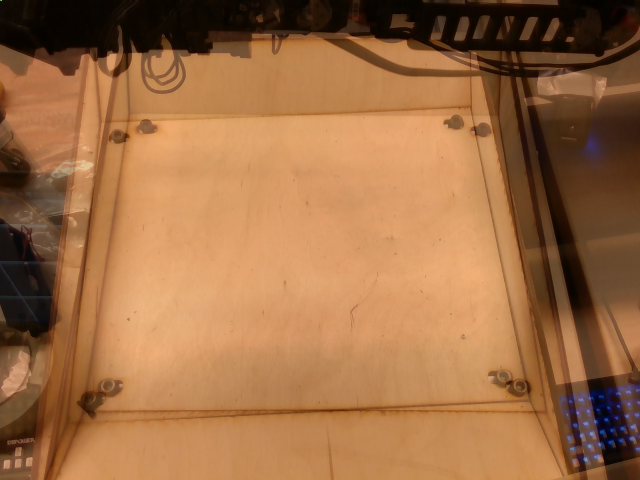} &
       \includegraphics[width=0.47\linewidth]{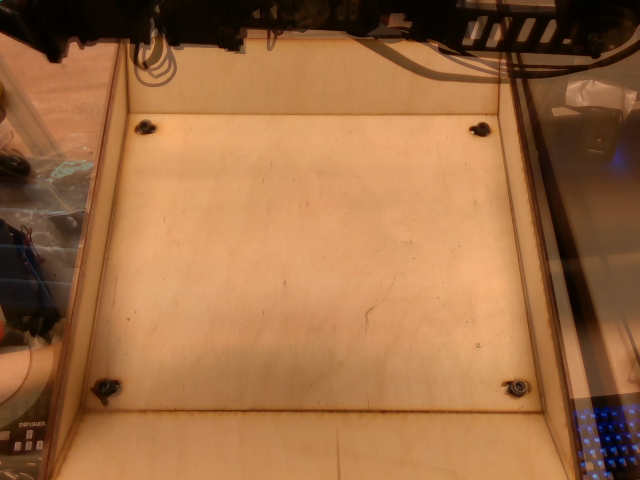}
  \end{tabular}
\caption{To calibrate REPLAB cells, we propose a protocol of finely adjusting the camera position until its camera image aligns nearly perfectly with that from the first REPLAB cell. Here, images from our two REPLAB cells are shown overlaid on top of each other before (left) and after (right) alignment.}
  \label{fig:calibration}
\end{figure}
In particular, for each cell after the first, we propose to finely adjust the camera position so that its view of its workspace and robot are aligned to those from the first cell camera. Fig~\ref{fig:calibration} shows this calibration protocol in action. While calibration from scratch is frequently time-consuming, this protocol enables simple calibration for all cells, and also helps ensure that all cells are near-identical in construction. We have applied this protocol in constructing our second REPLAB cell, and verified that it works in practice. Sec~\ref{sec:reproducibility} presents quantitative evidence for this. %

\subsection{Control Noise Compensation}

For all arm motions, we use ROS for inverse kinematics and planning with the \emph{MoveIt} package and execution through PID position control. While calibration noise is minimal, control noise is more difficult to avoid given that we use low-cost arms~\cite{deisenroth2011learning,gupta2018robot}.

In our setting, we found that control noise is primarily along the horizontal coordinates $(x, y)$. We tackle this using a simple approach. Since most desired grasping targets are near the cell floor, we first command the arm to move its end effector over a 5x5 grid on the floor and record the actual achieved positions of the end effector using the planner and controller described above. Comparing the target positions $p_i$ and achieved positions $q_i$, we fit a linear model $q=\alpha p+\beta$, where parameters $\alpha$ and $\beta$ are learned for each cell separately. %

Having calibrated the control noise, we can compensate for it by setting the target position to $p'=(p-\beta)/\alpha$. For our two cells, we set $\beta$ to 0 and $\alpha$ to 0.87 and 0.95. We find that this simple approach works well to eliminate most control noise. Combining camera calibration noise and residual control noise after compensation, the end-effector positions are within 2 cm of the target over the 35 x 40 cm workspace floor, and within 1 cm near the center. Qualitatively, this error falls within the tolerance that the grasping task (defined below) naturally permits.

\section{Supervised Learning for Grasping}

One of REPLAB's intended functions is to serve as a common platform for benchmarking control algorithms for robotic manipulation tasks. We now describe the template of a benchmark focused on arguably the longest studied manipulation task: grasping.

\subsection{Task Definition} %
Multiple objects are randomly scattered over the cell floor. Each grasp attempt may target any object in the workspace. All algorithms have access to a standard input-output interface. The RGBD image and raw point cloud from the camera are available as inputs. The RGB image, depth image, and point cloud are shown for a sample grasp attempt in Fig~\ref{fig:inputs}.  The algorithm output is required to be a fixed target grasp configuration.

We restrict the gripper to be oriented vertically.
This restriction is used in a number of prior works~\cite{levine2018learning,pinto2016supersizing,zeng2017robotic,calandra2017feeling} and significantly simplifies inverse kinematics and planning, since the arm is unlikely to collide with clutter during motion towards a grasp point.
A grasp is specified fully by a robot coordinate 3D point $(x, y, z)$, and a wrist angle $\theta$ for the parallel-jawed gripper. The arm is moved first to a position directly above the intended grasp, and then lowered to the correct grasp position. Once the target coordinates are achieved, the parallel-jaw gripper is closed, and the arm is moved into a preset standard configuration, with the gripper facing the camera, and held for two seconds. A successful grasp requires the object to stay in the gripper throughout this time. This protocol is common to all evaluated approaches. Grasp success detection is discussed in Sec~\ref{sec:data_collection}.

\subsection{Objects}
As pointed out in Sec~\ref{sec:related}, standard object sets for grasping have previously been proposed in~\cite{calli2015benchmarking,leitner2017acrv}. However, since these object sets were designed for much larger arms, we design new object sets for REPLAB --- a training set with over 100 objects of varying shapes and sizes, a ``seen object'' test set of 20 toys among the training objects, and an ``unseen object'' test set of 20 toys. Our objects are of varying shapes and sizes, with about 50\% hard plastic toys and 50\% soft toys. We specify shopping lists on the project page for reproducibility. %
Toys are picked so that there is at least one feasible stable grasp with the gripper. Some sample objects are shown in Fig~\ref{fig:objects}.

\subsection{Dataset and Data Collection}\label{sec:data_collection}

We have collected a dataset of over 92k randomly sampled grasps together with grasp success labels collected using two REPLAB cells in parallel, at the rate of about 2.5k grasps per day per cell with fewer than two interventions per cell per day on average. Roughly 23\% of grasp attempts during random data collection result in successful grasps. For each grasp attempt, the 3D point cloud returned by the camera is clustered using DBSCAN~\cite{ester1996density} to find objects, and a random cluster is selected. A grasp is sampled as follows: grasp coordinates $(x, y, z)$ are sampled with a small random perturbation from the center of the selected cluster. The grasp angle $\theta$ is sampled uniformly at random.

We collected the data across two cells under different illumination conditions and backgrounds. On cell 1, we collected approximately 44k samples under artifical room lighting. On cell 2, we collected another 44k samples near a window, with largely natural lighting. Finally, we collected an additional 4k samples on cell 1 under a more controlled setup, with powered light strips inside a fully enclosed cell. At test time, we evaluated in the latter two settings.

\subsection{Evaluation and Performance Metric}\label{sec:metrics}
Evaluation is done on an episodic bin-clearing task. At the start of an episode, 20 objects are scattered over the workspace floor using a fixed protocol: a box is filled with the objects, shaken, and inverted over the center of floor, similar to~\cite{pas2015using}.
Each episode consists of 60 grasp attempts. For each grasp attempt, 500 grasp candidates are evaluated from the neighborhood of each cluster returned by DBSCAN. In particular, we sample $(x,y,z)$ from points in the cluster and sample $\theta$ uniformly at random. Each successfully grasped object is automatically discarded to a clearing area outside the workspace, and one of the remaining objects must be picked at the next attempt. In rare cases when either clustering fails (number of objects is too low or too high), or there has been no successful grasp in 10 attempts, we sweep the arm over the workspace floor to perturb objects.
We report cumulative success rate (CSR) plots of the number of successfully grasped objects vs.~the number of grasp attempts. See Fig~\ref{fig:results} for an example. %

\section{Reinforcement Learning for Reaching}\label{sec:rl}
Many of the successes of reinforcement learning (RL) thus far have come in simulated domains. RL for robotic control remains challenging~\cite{kober2013reinforcement} especially with visual inputs~\cite{levine2016end,nair2018visual,andrychowicz2018learning}. We believe that as a reproducible hardware platform, REPLAB could accelerate research in this area by enabling shareable implementations, analogous to ALE~\cite{bellemare2013arcade} and Open AI Gym~\cite{gym} for simulated control tasks. RL has been reported to be extremely sensitive to factors of variation such as network architecture, hyperparameter choices, reward scaling, random seeds, environment specification, and specific choice of implementation~\cite{henderson2018deep}. We believe therefore that a real-world robotic control benchmark for RL is timely and important. Towards this, we implement and demonstrate a state-of-the-art off-the-shelf RL algorithm, TD3~\cite{fujimoto2018addressing}, on REPLAB for a basic control task, 3D point reaching, defined below. We train TD3 as implemented in RLkit~\cite{rlkit}.

\subsection{Task definition}
Reaching is arguably the most basic of manipulation skills and is a common setting for evaluating RL algorithms for robotic control. The reaching task requires controlling the arm so that its end-effector correctly reaches a target point. In our experiments, the target point is pre-specified. Joint angles are available as encoder readings of the servos, encapsulating six degrees of freedom. The 3D position (x, y, z) of the end-effector, a function of the joint angles, is also made available as part of the observation space for the RL algorithm. We provide a joint velocity control interface to the algorithm --- it must learn to control the six joint velocities over time to manipulate the end-effector to the target. We report performance in terms of distance from the target, as a function of training time. The reward at each step is the negative euclidean distance of the end-effector from the target.

\section{Plug-and-Play Software}

We aim to lower the barrier to entry into manipulation research not just by keeping REPLAB costs low, but also by emphasizing ease of use and reproducible algorithm implementations. In particular, all code is available in a Docker image that runs nearly out of the box on Ubuntu machines with very minimal setup, for quick reproducibility. The Docker image contains scripts for grasping (automatic data collection, grasp success annotation through the trained classifier, camera calibration, noisy control compensation, and evaluation) as well as reaching (TD3 algorithm implementation for reaching). For the grasping benchmark, it also includes REPLAB-specific implementations of several baselines for grasping, described in the next section. With this image, setting up an Ubuntu laptop to start collecting data on a REPLAB cell takes only a few minutes.  %

The importance of such plug-and-play implementations in accelerating research progress cannot be overstated, and we believe that this is one of the key advantages of a fully reproducible and standardized hardware platform. Going forward, we will invite and encourage authors of the leading approaches on REPLAB benchmarks to contribute their own implementations to include in future versions of the REPLAB open-source software packages. %

\begin{figure*}[t]
  \includegraphics[width=1\linewidth]{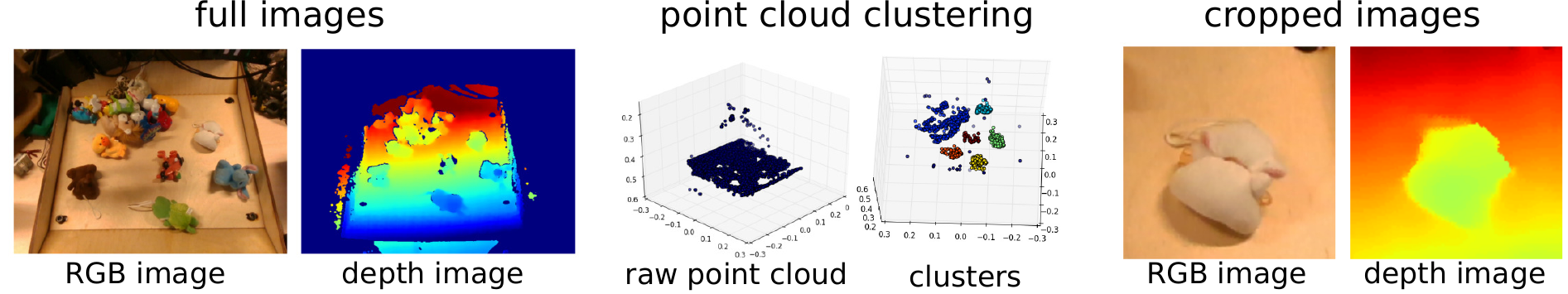}
  \caption{Raw and preprocessed RGB images, depth maps (blue is close, red is far), and pointclouds used in various grasping approaches. {\small\texttt{full-image}} operates directly on the raw RGB and depth images. {\small\texttt{random-xyz\straighttheta}}, {\small\texttt{random-\straighttheta}}, and \texttt{principal-axis} rely only on discovering clusters in the point cloud. Point cloud clustering is shown in the middle, where background points are removed from the point cloud before running DBSCAN. In this example, DBSCAN successfully gets the isolated objects and fails on the objects clumped together in the top left, detecting them as a single cluster. \texttt{cropped-image} uses input images cropped to the neighborhood of the candidate grasp (shown here to the right).}
  \label{fig:inputs}
\end{figure*}

\section{Experiments}~\label{sec:exp}

We now present experiments that aim to answer the following questions: (i) Is manipulation feasible on the low-cost REPLAB platform despite noisy control? (ii) How suitable are REPLAB cells to serve as foundations for a manipulation benchmark? In particular, does our grasping protocol produce consistent, reproducible evaluations across multiple REPLAB cells? (iii) What are the best-performing baseline approaches on the REPLAB evaluation protocol, and what can we learn by analyzing their performance? (iv) How well does a state-of-the-art reinforcement learning approach perform on a 3D reaching task on REPLAB?

\subsection{Grasping Approaches}

We implement and evaluate five grasping approaches in all. The first three are based on sampling near clusters detected in the point cloud: (i)  {\small\texttt{random-xyz\straighttheta}}: grasp angle $\theta$ is sampled uniformly at random, and grasp coordinates $(x,y,z)$ are perturbed with random uniform noise in a 4x4x2 cm region from each cluster center, (ii)  {\small\texttt{random-\straighttheta}}: Only $\theta$ is random, where $(x,y,z)$ is set to a cluster center, and (iii)  {\small\texttt{principal-axis}}: we find the major axis of a cluster by computing the largest eigenvector of the correlation matrix of $(x,y)$ coordinates of points in the cluster. A grasp is attempted along the perpendicular bisector of this axis. $z$ is fixed to the cluster center.

We evaluate two approaches based on training convolutional neural networks to predict the quality of a grasp in a given scene: (i)  {\small\texttt{full-image}}, based on~\cite{levine2018learning,calandra2017feeling}, takes two inputs: the full image of the workspace and the full $(x, y, z, \theta)$ parameterization of a candidate grasp, (ii)  {\small\texttt{cropped-image}}, based on~\cite{pinto2016supersizing} instead crops the input image around the $(x, y, z)$ position of the candidate grasp and predicts success or failure for each of 18 quantized $\theta$ bins. The inputs are schematically shown in Fig~\ref{fig:inputs}. These are both trained on the same set of 92k random grasps described in Sec~\ref{sec:data_collection}.

For testing on the robot, each baseline approach is provided with grasp candidates from which it picks one to execute.  {\small\texttt{full-image}} and  {\small\texttt{cropped-image}} evaluate the grasp quality of 512 grasp candidates per detected cluster, each parameterized by $(x, y, z, \theta)$ as described in Sec~\ref{sec:metrics}, before executing the highest quality grasp. {\small\texttt{random-\straighttheta}} and {\small\texttt{random-xyz\straighttheta}} select one cluster at random  before selecting grasps near that cluster center. For  {\small\texttt{principal-axis}}, we assign a confidence score to each cluster based on the ratio of the largest eigenvalue to the smallest, and select the cluster with the highest confidence.

In practice, selecting only the highest confidence grasp tends to lead to the arm getting stuck in a loop attempting the same unsuccessful grasp over and over. To prevent this, we sample from top-5 grasp candidates for the learning approaches and the top 5 clusters for  {\small\texttt{principal-axis}}.

\begin{figure*}
  \includegraphics[width=0.33\linewidth,trim=0.2cm 0 0.5cm 0.3cm, clip]{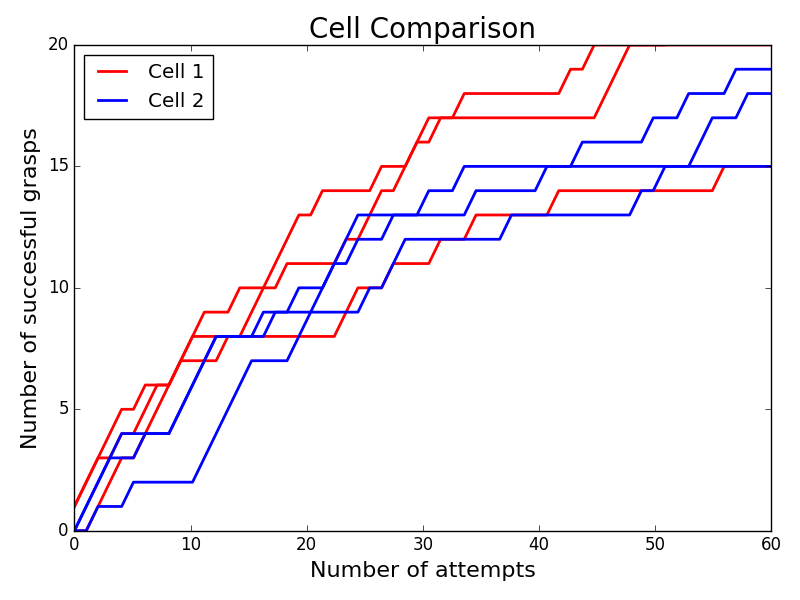}
  \includegraphics[width=0.33\linewidth,trim=0.2cm 0 0.5cm 0.3cm, clip]{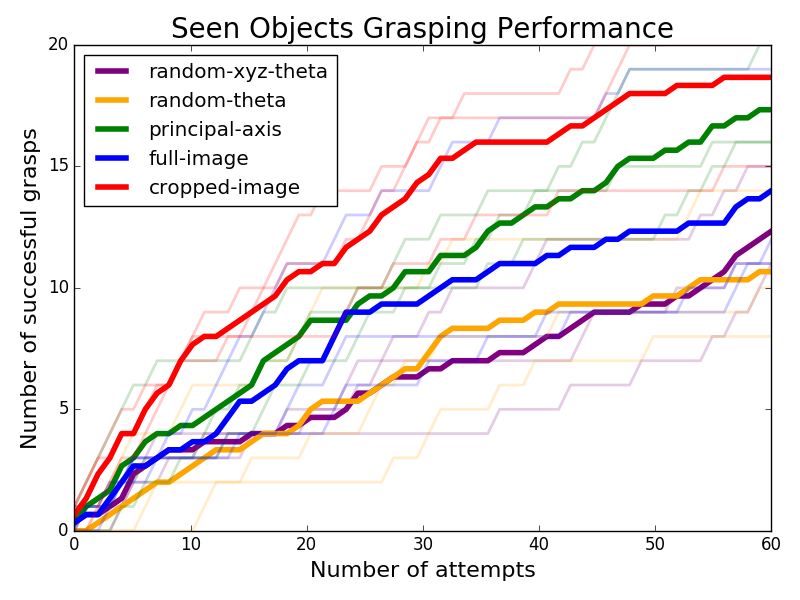}
  \includegraphics[width=0.33\linewidth,trim=0.2cm 0 0.5cm 0.3cm, clip]{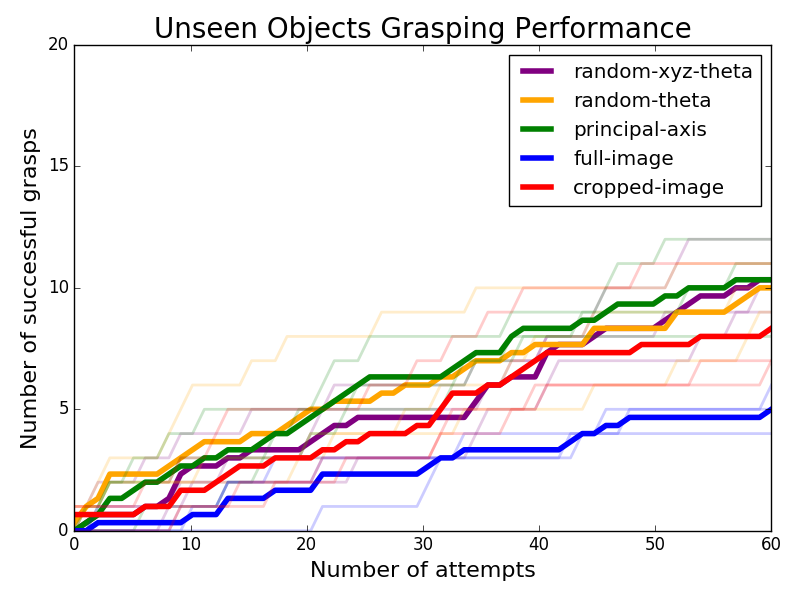}
  \caption{\textbf{(Left)} Cumulative Success Rate (CSR) plots from three runs on each REPLAB cell for the  {\footnotesize\texttt{principal-axis}} baseline. Quantitatively and qualitatively, we observe very similar behavior on both cells with the same model, verifying the reproducibility of the REPLAB platform. \textbf{(Middle)} Cumulative Success Rate (CSR) plots for all baselines on seen objects. The mean over three runs is plotted in the thick curves, while the faded lines show individual runs for each method. \textbf{(Right)} Cumulative Success Rate (CSR) plots for all baselines on unseen objects. The  mean  over  three  runs  is  plotted  in  thick  curves,  while the faded lines show individual runs for each method.}
  \label{fig:results}
  \label{fig:cell_compare}
  \label{fig:seen_csr}
  \label{fig:unseen_csr}
\end{figure*}

\subsection{Reproducibility}\label{sec:reproducibility}
We have taken care in the design of REPLAB cells to make it possible to construct near-identical replicas. A reproducible hardware platform is key to establishing reproducible evaluation procedures, which is the primary aim of REPLAB. Evaluations of control algorithms should produce similar results on all REPLAB cells.

With this in mind, we have proposed a calibration protocol in Sec~\ref{sec:calibration} so that two REPLAB cells should in theory share the exact same calibration matrix $C$ mapping camera coordinates to robot coordinates. We have constructed two REPLAB cells using this procedure---the second cell inherits the calibration matrix computed for the first cell. To evaluate whether the cells are indeed constructed near-identically, we collect a small dataset of 25 corresponding camera and robot coordinate points $p_{cam}$ and $p_{arm}$ in each cell using a checkerboard (similar to correspondences used in calibration). We then measure the average calibration errors $\|Cp_{cam}-p_{arm}\|_2$ for each arm---if the cells are indeed identical, calibration errors should be similar for both cells. For the first and second cell respectively, the errors are 0.87 cm and 1.52 cm. Given a gripper of width 3 cm, this difference is tolerable and in practice leads to the same grasping behavior, as we show below.

To measure reproducibility in the context of grasping evaluation, we evaluate the  {\small\texttt{cropped-image}} grasping baseline over three runs on each cell separately. Fig~\ref{fig:cell_compare} shows cumulative success rate (CSR) plots. As seen here, the population of CSR curves is very closely matched across the two cells. %

Note that, as described in Sec~\ref{sec:data_collection}, we only collected 4k samples ($<$ 1.5 days) in the testing condition of Cell 1. As Fig~\ref{fig:cell_compare} shows, this suffices to produce the nearly identical results to Cell 2, where we collected over 40k training samples. We expect therefore, that for evaluation under unseen conditions (such as illuminations or backgrounds that are not present in our training dataset), it will suffice to finetune our models, shared on the project website, with 4k training samples from the new setting. Our Docker image includes code for collecting this additional data, annotating it, and retraining models. Analytical approaches such as {\small\texttt{principal-axis}} do not require any training data and work out-of-the-box under previously unseen conditions.

\subsection{Performance of Grasping Baselines}
We now evaluate all five baselines on our grasping task.  %
First, our learning-based baselines {\small\texttt{cropped-image}} and {\small\texttt{full-image}}, trained on our dataset of random grasps, yield 64.8\% and 62.8\% accuracy respectively on a balanced, held-out validation dataset of 10k sampled grasps.

Moving to on-robot evaluation, Fig~\ref{fig:seen_csr} (middle) shows the cumulative success rate (CSR) curve on seen objects.  {\small\texttt{cropped-image}} clears all 20 test objects two out of three times, emerging as the leading approach, followed by {\small\texttt{principal-axis}}, and {\small\texttt{full-image}}. We believe that the advantage of  {\small\texttt{cropped-image}} over {\small\texttt{full-image}} comes from preprocessing the image inputs to focus on the region of interest. In contrast, {\small\texttt{full-image}} must \emph{learn} this association between grasp parameters and image locations with only grasp success/failure as supervision. {\small\texttt{full-image}} also requires a larger network to process its larger inputs, making it more prone to overfitting. We expect that this gap in performance will fall as the size of the grasping dataset increases. %
{\small\texttt{principal-axis}} relies heavily on discovering objects through clustering, and has high variance early in evaluation runs when objects clump together.

Fig~\ref{fig:unseen_csr} (right) shows the CSR curves for unseen objects that were not encountered during training. By design, our unseen objects are significantly more complex shapes than were seen at training time, as shown in Fig~\ref{fig:objects}. Unsurprisingly, all methods perform worse on this set. {\small\texttt{principal-axis}} explicitly relies on objects having simple ellipsoidal geometries, and struggles to handle these more complex shapes. The learning-based approaches {\small\texttt{cropped-image}} and {\small\texttt{full-image}} are limited in their ability to generalize to these objects by the fact that training data was collected using only objects with relatively simple geometries---we found that a training set of simple objects was important to ensure sufficient success rate (about 23\%) at data collection time so that there were enough successful grasps in training data. We expect that learning-based methods will benefit from a curriculum-based approach for collecting a larger dataset, where the current best policy may be deployed (with some exploration) to collect data on increasingly more difficult objects. %

Note that over all approaches and all trials reported in Fig~\ref{fig:results}, the fastest clearance ({\small\texttt{pinto2016}}) still takes over 40 attempts to clear 20 objects. Together with the performance on the  unseen object set, this is a good sign for a REPLAB-based grasping benchmark; reasonable baselines still have plenty of room for improvement.

Finally, we perform a data ablation study on $\small{\texttt{cropped-image}}$, evaluating only held-out validation accuracy on the seen object grasps. With 5k, 10k, 20k, 40k, and 80k training grasps, accuracies are 57.8\%, 56.3\%, 62.9\%, 64.4\%, and 64.8\% respectively. The diminishing returns suggest that our dataset is large enough to train this model for the seen objects. However, we expect that more data would benefit larger models, as well as the ability to generalize to unseen objects. Our entire dataset is available on the project webpage, and we plan to further grow this dataset before releasing a grasping benchmark. Towards this, we invite dataset contributions from the robotics community. Our software packages include data collection and annotation scripts to make it easy for robotics researchers to make such contributions towards the REPLAB grasping benchmark.

\subsection{Performance of RL-Based Reaching}\label{sec:rl_results}
\begin{figure}
  \centering
  \includegraphics[width=0.7\linewidth]{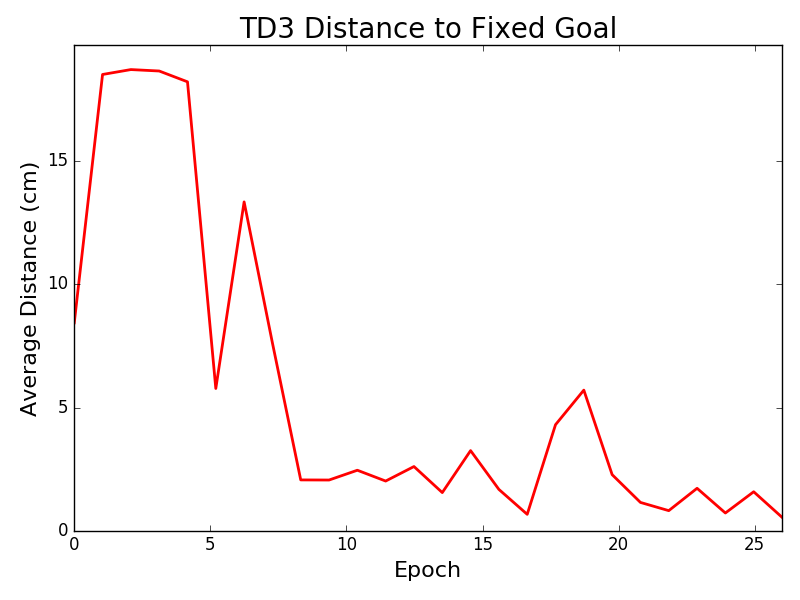}
  \caption{Plot of average distance of end-effector from the target 3D point (in cm) after each epoch of reinforcement learning. Within 25 epochs ($\sim$ 1 day of training), reaching errors fall to under 1 cm, or learning to control joint velocities through the TD3 reinforcement learning algorithm.}
  \label{fig:rl_plot}
\end{figure}

Finally, we evaluate the performance of the TD3~\cite{fujimoto2018addressing} reinforcement learning algorithm for 3D point reaching, as described in Sec~\ref{sec:rl_results}. Fig~\ref{fig:rl_plot} shows the reaching error over training time. Within 25 epochs of training (about 24 hours in total), reaching errors converge to less than 1 cm, demonstrating the feasibility of using REPLAB as a platform for reinforcement learning research. Being easily scalable to clusters of many cells, REPLAB holds promise for enabling RL approaches for complex tasks that require higher sample complexity, through parallelized training of RL algorithms.

\section{Future Work}

We have proposed a fully standardized hardware stack on which to develop reproducible evaluation procedures for manipulation tasks. To illustrate the use of such a platform, we have provided the template of a grasping benchmark, and also illustrated shareable robotic reinforcement learning implementation for a reaching task. One immediate shortcoming with the current platform in terms of its widespread adoptability is the reliance on a specific robot arm supplier. We plan to address this in future versions of the REPLAB platform through a 3D-printable arm design.
We also plan to build upon this foundation by
building and releasing Gym environments for REPLAB simulation experiments, and implementing more grasping approaches and RL algorithms on REPLAB.

As a standardized hardware platform, REPLAB aims to enable sharing both:
(i) implementations of control algorithms, such as our grasping and RL algorithms, and
(ii) pretrained models that work out of the box, to enable easy benchmarking and reproducibility. %
While we already share algorithm implementations on the project website, our pretrained grasping models currently require additional finetuning (about 4k new training samples, taking about 1.5 days to collect and label) to work on new REPLAB cells, under unseen lighting conditions and backgrounds. We are working towards developing a more diverse training dataset for a grasping challenge that would allow direct generalization to new conditions.

We plan to also release open-source code for robotic control approaches such as visual servoing, video prediction-based model predictive control, and other reinforcement learning approaches for the REPLAB platform. We invite other datasets, task definitions/challenges, and software contributions from the robotics research community.

\section*{ACKNOWLEDGMENTS}
We thank Dian Chen and Fred Shentu for their help in early stages of the project, Ashvin Nair, Jeff Mahler, David Gealy, Roberto Calandra, and Jitendra Malik for helpful discussions, Mohammad Kamil Shams and Alan Downing for participating in verifying the reproducibility of REPLAB and providing useful feedback, and Trossen Robotics for their assistance with the WidowX arm.

\bibliographystyle{IEEEtran}
\bibliography{refs}

\end{document}